%% file: 0_AERO_2026.tex
\documentclass[twocolumn,letterpaper]{IEEEAerospaceCLS}  

\usepackage[]{graphicx}    
\usepackage[colorlinks=true, urlcolor=black, linkcolor=black, citecolor=black]{hyperref} 
\newcommand{\ignore}[1]{}  
\input{1_preamble.tex}

\begin{document}
\title{SLAP: Slapband-based Autonomous Perching Drone with Failure Recovery for Vertical Tree Trunks }

\author{%
Julia Di\\
Stanford University\\
420 Panama Mall\\
Stanford, CA 94305\\
juliadi@alumni.stanford.edu
\and 
Kenneth A. W. Hoffmann\\
Stanford University\\
420 Panama Mall\\
Stanford, CA 94305\\
khffmnn@alumni.stanford.edu
\and
Tony G. Chen\\
Georgia Institute of Technology\\
801 Ferst Drive\\
Atlanta, GA 30332\\
tonygchen@gatech.edu
\and
Tian-Ao Ren\\
Stanford University\\
420 Panama Mall\\
Stanford, CA 94305\\
tianao@stanford.edu
\and
Mark Cutkosky\\
Stanford University\\
420 Panama Mall\\
Stanford, CA 94305\\
cutkosky@stanford.edu
\thanks{\footnotesize 979-8-3315-7360-7/26/$\$31.00$ \copyright2026 IEEE}              
}

\maketitle

\thispagestyle{plain}
\pagestyle{plain}

\maketitle

\thispagestyle{plain}
\pagestyle{plain}

\begin{abstract}
Perching allows unmanned aerial vehicles (UAVs) to reduce energy consumption, remain anchored for surface sampling operations, or stably survey their surroundings. Previous efforts for perching on vertical surfaces have predominantly focused on lightweight mechanical design solutions with relatively scant system-level integration. Furthermore, perching strategies for vertical surfaces commonly require high-speed, aggressive landing operations that are dangerous for a surveyor drone with sensitive electronics onboard. This work presents the preliminary investigation of a perching approach suitable for larger drones that both gently perches on vertical tree trunks and reacts and recovers from perch failures. The system in this work, called SLAP, consists of vision-based perch site detector, an IMU (inertial-measurement-unit)-based perch failure detector, an attitude controller for soft perching, an optical close-range detection system, and a fast active elastic gripper with microspines made from commercially-available slapbands. We validated this approach on a modified 1.2~kg commercial quadrotor with component and system analysis. Initial human-in-the-loop autonomous indoor flight experiments achieved a 75\% perch success rate on a real oak tree segment across 20 flights, and 100\% perch failure recovery across 2 flights with induced failures. Videos and code of flight tests can be found at our project website: \url{website-released-on-acceptance}.
\end{abstract}

\tableofcontents

\section{Introduction}

\begin{figure}[!t]
\centering
\includegraphics[width=1\columnwidth]{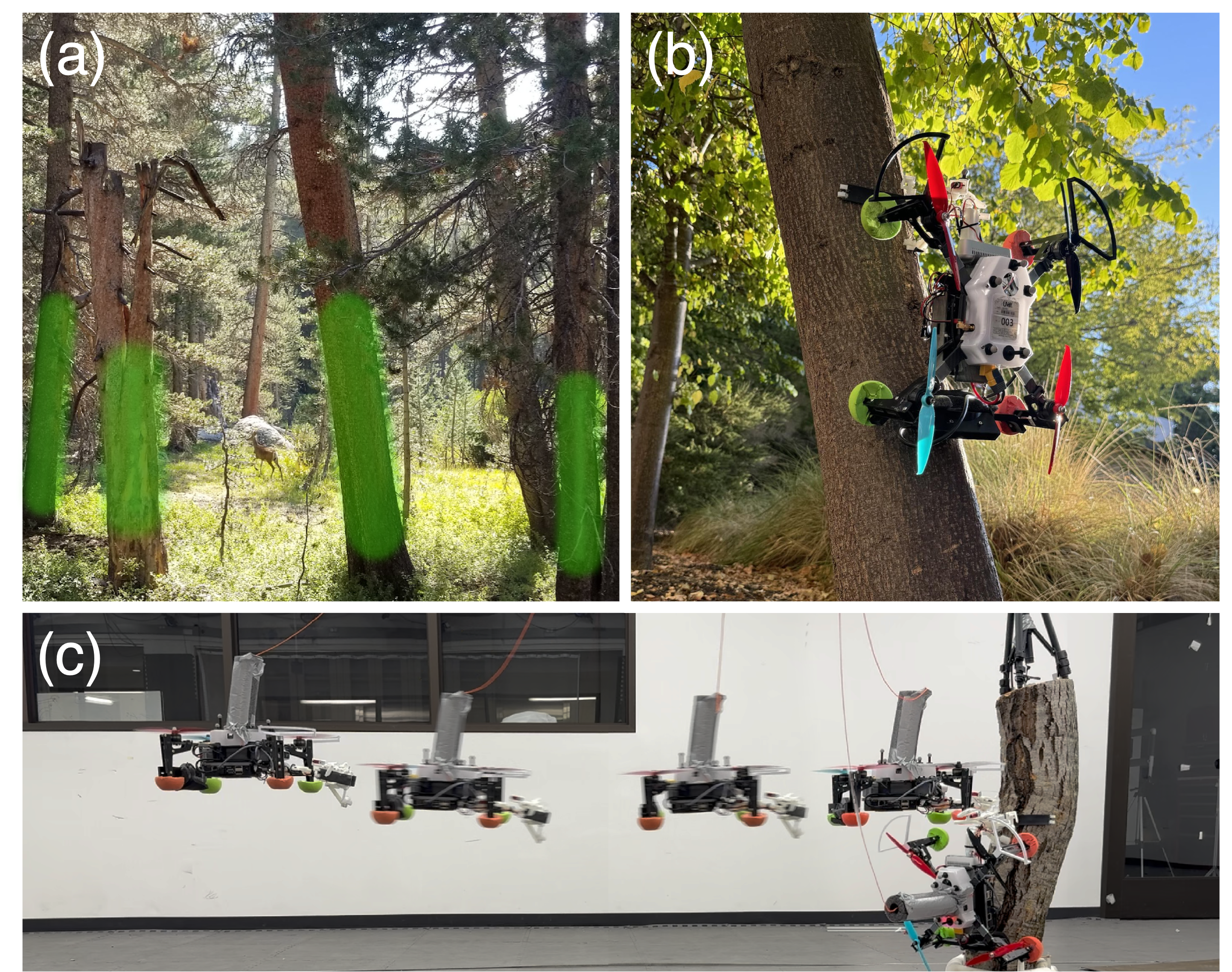}
\captionsetup{justification=justified}
\caption{(a) In mature forests, vertically-oriented tree trunks as highlighted in green are accessible areas for large drones to perch. (Photo taken by J. Di in Inyo National Forest.) Perching payload-capable drones could enable environmental monitoring of the surroundings. (b) A concept photograph of the perched drone system posed on a basswood tree. (c) A composite photo of a human-in-the-loop autonomous perching sequence of \SI{1}{\kilo\gram}-class quadrotor on a real oak tree segment. The system identifies the perch site, flies a polynomial trajectory, and perches autonomously. The human verifies the perch site identified by the system before engaging perching.}
\label{fig:main}
\end{figure}

In nature, the ability to perch enables animals to hunt, monitor surroundings, and rest between flights. Similarly, aerial robots like quadrotors are unmatched for surveying and inspection tasks. When equipped with perching capabilities, quadrotors could quietly monitor wildlife, sample from tree surfaces, and extend their operational time through solar energy harvesting~\cite{meng2022aerial,roderick2017touchdown}. 

Inspired by the avian perching abilities, many studies consequently have developed mechanical designs and control strategies for aerial robot perching. Existing work for vertical surfaces often execute pitch-up or collision maneuvers to reduce impact energy~\cite{mellinger2010control,lussier2011landing,pope2016multimodal,kalantari2015autonomous}. Although these maneuvers are effective for lightweight drones in relatively open areas, they become more difficult or even dangerous to execute for larger, heavier drones with payloads.

To avoid risky pitch-up operations, other studies have investigated perching on horizontally-oriented surfaces like branches and rooftops. In these scenarios, a drone may perch by sitting on a target, where the grasp is often accomplished through passive mechanical solutions such as avian-inspired claws, wrapping arms, or simply adhesives that provide a hold onto a surface~\cite{doyle2012avian, roderick2021bird,askari2024crash}. While horizontal surfaces like branches may be better suited for bearing heavier drones, they may not be easily accessible or provide only obstructed views of the surroundings. In mature arboreal environments, vertical tree trunks instead present an attractive perch site alternative. 

In this work, we propose a method for gently perching payload-capable mulitrotor systems on vertically-oriented tree surfaces, and report the preliminary results from flight tests. Practical arboreal use cases include wildlife monitoring, biodiversity mapping, and surface sampling; perching on man-made vertical poles, such as wooden utility and electrical poles, are also applicable.  Motivated similarly, Askari et al. recently introduced a passive wing morphing design that allows aerial gliders to crash into vertical poles and wrap its wings in an enveloping grasp~\cite{askari2024crash}. However, for drones, we argue that high-speed approaches are undesirable and instead propose an autonomous perching approach for vertical surfaces that does not require aggressive flight conditions. Similarly, Li et al. introduced the Treecreeper Drone, a passively triggered perching mechanism for vertical tree trunks~\cite{li2025treecreeper}, but the work does not address active control or recovery from perch failure. In contrast, our system emphasizes active grasping, failure detection and recovery, and low-velocity (gentle) perching on payload-capable drones, thereby enabling safer and more robust operation in real-world arboreal settings. Our system consists of a vision-based perch site detector, an IMU-based perch failure detector, an attitude controller for soft perching, an optical close-range detection system, a mechanical pivot point, and an active elastic gripper that exploits both compliant wrapping and microspine technology for surface adhesion. 

As shown in \Cref{fig:main}, the flight-to-perch system is validated on a \SI{1.2}{\kilo\gram} Uvify IFO-S quadrotor in human-in-the-loop (HITL) autonomous experiments on an oak tree segment with localization provided from a motion capture system. Human input is used as a safety check for initiating the perch site detector and initiating the perching sequence. The perch site detection, trajectory generation, flight, perching sequence, and perch failure recovery are done autonomously. This work also reports the system development process that led to flight demonstrations. 

The contributions of this work are:
\begin{itemize}
    \item Fast active gripper design using bistable elastic bands and microspines for vertical tree trunk perching, enabling slow approach flight speeds
    \item System engineering approach and integration of perception, planner, and a gentle perch controller for tree trunk perching
    \item Design of an IMU-based perch failure recovery routine
    \item Human-in-the-loop autonomous indoor system demonstrations of perching and perch failure recovery on oak tree trunk
\end{itemize}

\section{Related Work}
In recent years, new methods for aerial perching robots have been proposed and extensively reviewed~\cite{meng2022aerial,roderick2017touchdown}. To counteract the dynamic forces involved for perching, astriction or grasping technologies are required. Many methods therefore focused on proposing new mechanical designs for securely attaching to the surface. Relevant to this work are earlier efforts using microspine grippers for attachment to rough surfaces~\cite{lussier2010landing,lussier2011landing,pope2016multimodal,kovavc2009perching}. However, these earlier microspine grippers have only been demonstrated on lightweight platforms in the hundreds of grams. 

This work specifically focuses on heavier UAVs for forest environments, where perching enables scientific monitoring and analysis. For arboreal perching of \SI{1}{\kilo\gram}+ platforms, similar approaches that have been successful include microspine-enabled grapples~\cite{nguyen2019passively} and avian-inspired passive claw grippers~\cite{ doyle2012avian, roderick2021bird}. Those previous designs have only been demonstrated manually and on only horizontally oriented cylindrical structures like branches.

For perching onto trees, vertically-oriented trunks are generally more accessible to aerial platforms as they are  largely free of occluding branches and leaves, but they complicate the control dynamics of perching. To safely perch on a surface, the UAV must store and/or dissipate kinetic energy at the end of the perching sequence. Previous efforts on lightweight platforms have accomplished this by entering high angles of attack and stall conditions prior to contact~\cite{kovavc2009perching,lussier2011landing,pope2016multimodal,moore2012control,askari2024crash}. These aggressive flight trajectories are risky, particularly for larger drones with significant payloads, and induce target loss and visual motion blur that could lead to perch failure. Furthermore, it leaves the UAV vulnerable upon takeoff from the perch site.

Instead, this work utilizes a forward-mounted perching mechanism designed to simplify the control requirements for vertical surface perching. While this approach introduces additional forward mass, it reduces the complexity of the control problem for vertical-surface perching compared to mechanisms mounted on the undersides. For heavier drones, the additional moment arm when perching must also be counteracted, which in this work we accomplish with a mechanical pivot point. This mechanical pivot is similar to the approach introduced by Li et al.~\cite{li2025treecreeper}; however the passive triggering design in their work has strict approach constraints and lacks analysis of the failure recovery of the system. In contrast, the gripper in this work uses an active perching mechanism with two bistable slap band bracelets to achieve rapid closure and intrinsic wrapping in addition to spine-based adhesion. Zheng et al. recently proposed an active gripper solution that also uses tape springs for manually perching both micro-UAVs and \SI{1}{\kilo\gram} UAVs onto horizontally-oriented branches~\cite{zheng2024albero}, but without microspines and demonstrated only for horizontal surfaces.

Finally, most perching flight demonstrations to date have been entirely manual and in controlled laboratory environments, but steady progress is being made towards system autonomy in realistic settings. Hang et al. had demonstrated a vision-based perching and resting drone for horizontal surfaces, though in an indoor motion capture environment~\cite{hang2019perching}. Zufferey et al. showed an ornithopter autonomously detecting and flying to a horizontal cylindrical structure in an indoor motion capture environment~\cite{zufferey2022ornithopters}. Perching efforts outdoors have usually been teleoperated, with a recent work in integrating autonomy into portions of the perching sequence of a force-sensorized sticky cage for horizontal branches~\cite{aucone2023drone}. Promising system-level integration has been demonstrated in related applications, such as outdoor powerline inspection and manipulation conducted by the multi-university AERIAL-CORE consortium~\cite{ollero2024aerial}, as well as outdoor high-speed (\SI{2}{\meter/\second}) aerial grasping of objects from horizontal platforms~\cite{ubellacker2024high}. However, for vertical surfaces, such integration to date has only been shown on lightweight platforms for aggressive pitch-up maneuvers in indoor motion capture laboratories~\cite{thomas2016aggressive,mao2023robust} or in passive systems like Treecreeper flown manually~\cite{li2025treecreeper}. Because falling from a vertical surface risks serious damage to the vehicle, detection and recovery of perch failures is a critical part of the system autonomy. Previous work has demonstrated that onboard accelerometer data is capable of discriminating perch failure for a \SI{20}{\gram} Crazyflie, but not for a significantly larger vehicle that may have different behavior~\cite{jiang2015perching}.

\section{Robot Design Approach}

To de-risk vertical surface perching for larger autonomous drones, this work sought to minimize the impact risk to the drone during the perching sequence. To accomplish this, we made two key system design choices: (i) elimination of the pitch-up maneuver by choosing a system design that supported forward flight only to perch, and (ii) design of the perching system for near \SI{0}{\meter/\second} impact velocity at the target.
\begin{figure}[!t]
\centering
\includegraphics[width=1\columnwidth, trim=0 20 20 0, clip]{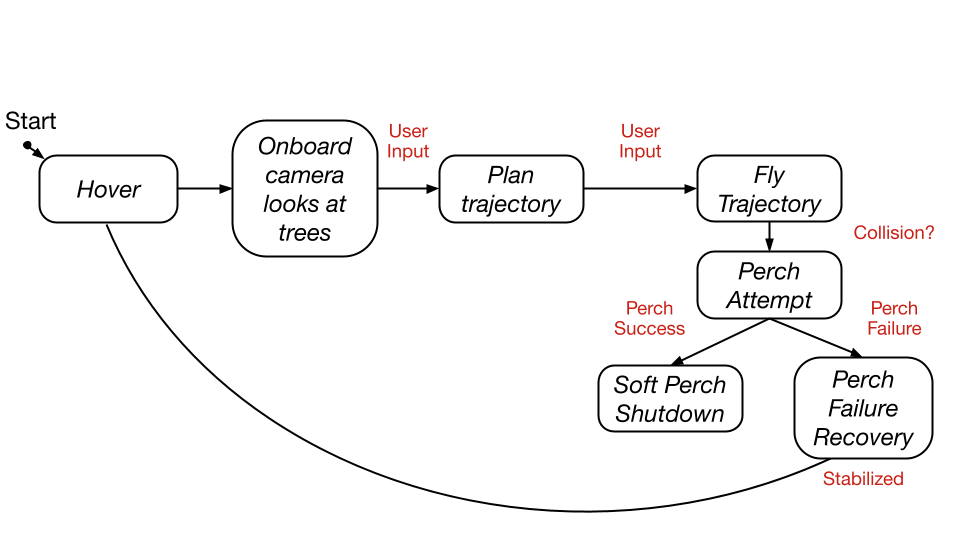}
\caption{Autonomy state diagram. The system has human-in-the-loop check-ins to verify tree detection and to confirm that the system should start the perching maneuver. In the event of a perch failure, the system detects free fall and stabilizes itself in a hover at a safe distance from the ground and the tree.}
\label{fig:systemoverview}
\end{figure}

To understand the design tradeoff from (ii), it is helpful to use the concept of the sufficiency region, which defines the envelope of design parameters for a successful grasp~\cite{roderick2021bird, chen2022aerial}. For grippers designed for impact with objects, we examine the sufficiency region in velocity space. Generally, passively-triggered grippers require some non-zero relative velocity between drone and target. Previous literature has shown a minimum required velocity around \SI{1}{\meter/\second}~\cite{chen2022aerial} and \SI{2}{\meter/\second}~\cite{li2025treecreeper} for different gripper designs. It could be too risky for a payload-bearing drone system to impact a tree at those speeds. Instead, to guarantee a lowered relative velocity requirement, we choose to actively control the gripper through the use of a time-of-flight (ToF) sensor and servo motor. This shifts the velocity sufficiency region to the left, such that the incoming velocity can be close to \SI{0}{\meter/\second}. However, the extra electronics incurs a mass cost. The selected Uvify IFO-S platform had margin in the mass budget for an active gripping solution, but this design tradeoff is not optimal for less payload-capable quadcopters. 

Additional objectives include to (i) minimize overall weight and inertia, (ii) maximize compensation for drone flight imprecision and (iii) impose as small as possible torques through the microspines during the perching sequence so that they do not rotate off the surface.

\subsection{Active Gripper Design and Avionics}

\begin{figure}[!t]
\centering
\includegraphics[width=1\columnwidth]{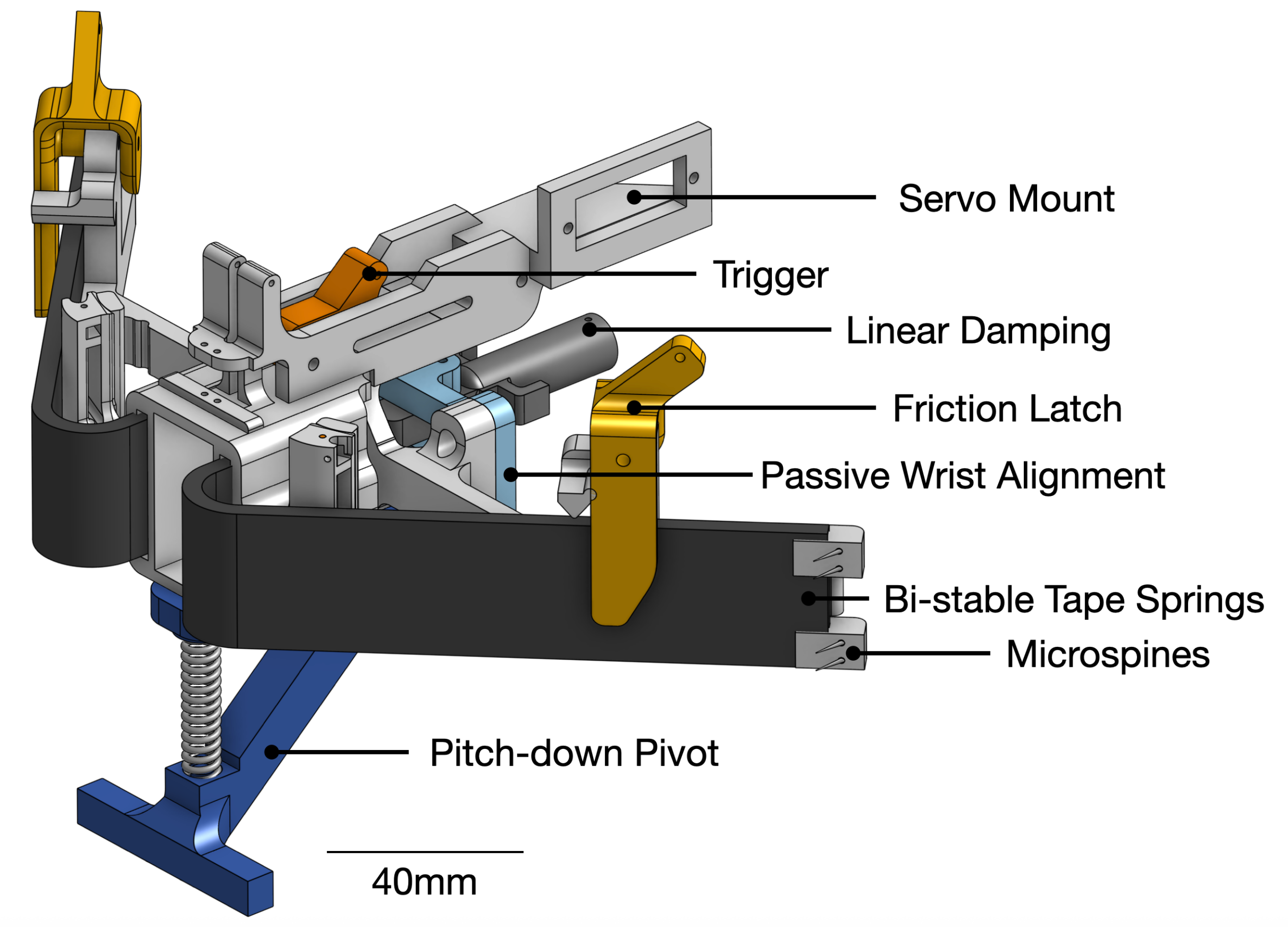}
\caption{CAD rendering of the gripper design. A single servo is used to trigger the two friction latches through two tendons, and in turn, the latches release two bi-stable tape springs with microspines mounted at the tip. }
\label{fig:gripper}
\end{figure}

The CAD rendering of the gripper mechanism used in this work is shown in Fig.~\ref{fig:gripper}. The gripper consists of four main subsystems: (i) grasp mechanism, (ii) trigger mechanism, (iii) wrist with built-in compliance, and (iv) a pitch-down pivot. Together, these elements enable rapid closure and secure adhesion to the tree bark. 

\subsubsection{Grasp Mechanism}

An active gripper necessarily incurs an additional grasp time delay compared to a mechanically passive gripper. To minimize grasp time as much as possible, the grasp mechanism utilizes two bistable tape springs modified from commercially available slap band bracelets (brand BRANDWINLITE BW-SB05-BK) after an initial exploration for spring stiffness and curvature. Each band is stored in the energized state until triggered, at which point the bands snap around the tree trunk within \SI{6}{\milli\second}.

To support a \SI{1.2}{\kilo\gram} mass on a tree, there are four microspines attached at the end of each bistable tape spring. The microspines are fabricated from organ needles, with the eyelets filled with epoxy to help prevent breakage. A pair of microspines are glued into a 3D-printed tile with cyanoacrylate adhesive, and two tiles are glued to the ends of the band.

Unlike rocks, tree bark is generally soft enough for a spine to penetrate. Because kinetic energy is proportional to the square of velocity, when the slap bands spring quickly while grasping, they have enough kinetic energy for the microspines to create their own asperities. This penetration ensures a secure grasp despite the lack of load sharing between the microspines. In addition to providing the initial preload for spine penetration, the slap bands also provide slight internal force for the microspines to stay engaged, as discussed in Appendix~\ref{appendix:gripperfea}. 
The gripper was 3D-printed on a Formlabs Form 3+ resin printer using Rigid 4k material and weighs only \SI{125}{\gram}.

\subsubsection{Trigger Mechanism}
In order to trigger the grasping sequence, a single servo (MKS HS75K) is used to pull an active trigger. Four music-wire extension springs with a spring constant of \SI{105}{\newton/\meter} are stretched between two dowel pins that are \SI{30}{\milli\meter} apart to provide \SI{3.15}{\newton} of force. When the active trigger is pulled, one of the dowel pins is released, thereby pulling on two tendons made from high-strength fishing line (``Power Pro Spectra 40lbs (178N)"). Each tendon internally routes to the corner of a friction latch, which pivots about a pin joint when pulled, releasing the stored tape springs simultaneously. The trigger requires manual reset between trials. 

A time-of-flight sensor (VL53L1X) inset between the two bands measures the distance to the tree, and electronically signals the servo to trigger once a threshold distance is reached.

\subsubsection{Wrist with Compliance and Suspension}
The gripper mechanism is designed such that at impact with the perch site target, mechanical compliance reduces the load on the drone and compensates for angular imprecision with the target. Two stabilizing rubber bands with a spring constant of approximately \SI{30}{\newton/\meter} are tied to the two sides of the structure to provide yaw angular compliance. This compliance helps compensate for angular imprecision with the target perch site by allowing the gripper to rotate, and absorbs some kinetic energy at impact.

The gripper suspension is made from two cylindrical tubes that slide within each other with a linear spring, as shown in Fig.~\ref{fig:gripper}. The front tube can slide into the back tube, with the distance set by a dowel pin. This linear compliance also helps to absorb the kinetic energy of impact.

\subsubsection{Pitch-down Pivot}
After a successful grasp, the elbow joint allows the drone to pitch down by up to 120 degrees, with a spring-loaded pivot at the base. This is necessary to provide an additional contact point against the tree during the pitch-down maneuver, preventing excess moments being applied to the microspines. The spring-loaded brace stows passively while resting on a flat surface.

\subsection{Gentle Perching}

\begin{figure*}[!t]
\centering
\includegraphics[width=\textwidth]{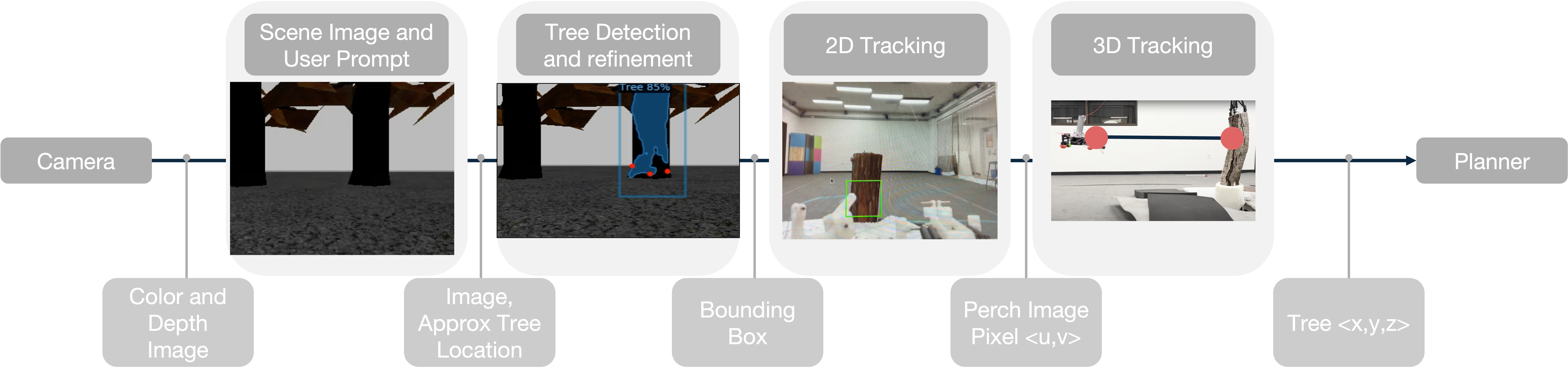}
\caption{Vision pipeline overview. The raw RGBD camera images are displayed to the user, who prompts the tree search. The tree detector then detects and displays the bounding box and keypoints of a suitable perch site, which then are continuously tracked by the drone during flight and fed to the planner.}
\label{fig:visionpipeline}
\end{figure*}

To prevent the microspines from slipping off an asperity, it is desirable to minimize the torques transmitted to the microspines. There needs to be a brace to counteract the downwards force of gravity after the drone has grasped the tree. Otherwise, the spines need to provide the opposing moment about which the drone can pivot. We accomplish this with the mechanical brace on the gripper, as well as a tail. 


One concern with the tail is that when the drone powers down, the impact with the tail on the tree could cause the microspines to slip off. Indeed, this happened during initial testing. This motivated the inclusion of a gentle perch controller that tuned the motor thrust based on drone attitude. The benefit of this design is after perching and successful grasp, the system become highly constrained, with essentially a single degree of freedom controlled by the collective thrust of the drone. We empirically determined the length of time and rate of decrease through testing, and found that decreasing from hover thrust to zero thrust over 4 seconds at \SI{100}{\hertz} worked well.

\subsection{Robot Onboard Autonomy}

\subsubsection{Vision-based Perch Site Detection} To detect the perching site, the detector uses an onboard Intel Realsense D435 RGBD camera that was rigidly mounted to the front of the drone. The camera outputs two images, a color frame and an aligned depth frame, in an ideal range from 0.3 to \SI{3}{\meter}. Because the images are aligned, each color pixel has a corresponding depth. In order to detect the tree mask, we use a deep-learned model based on PercepTree, a previously published CNN-based model for tree detection in forests~\cite{grondin2022tree}. 

The perception system has to be resilient to the perching hardware partially occluding the
onboard camera. This is possible by masking the regions of the image in which the gripper obscures the field of view because this is static throughout flight. The trade-off here is that the depth image is significantly occluded, and requires design iteration to balance the overall arrangement of the RGB-D camera with respect to the grasper. 

Based on the tree mask, we can then select the final perch site based on a number of factors: (i) the diameter of the tree section, (ii) the tree bark texture which can be learned and identified based on tree species~\cite{othmani2014tree}, and (iii) the contour of the tree. 


The (ii) tree bark texture is important for two reasons. First, as mentioned previously, microspines perform reliably on soft bark because the slap band spring has enough kinetic energy to penetrate the bark. Second, in initial tests, we found that older and larger trees tended to have flakier bark with deep fissures, which could cause perch failures because the surface would peel. As a result, we aimed for trees with surface textures that were uniform and therefore more predictable in performance. We did not implement a specific tree species identifier for texture analysis, but this could be done with a deep learning approach. In this work, we only tested the perception system on oak and basswood trees, which have softer bark. However, other trees can be identified based on their 3D bark texture~\cite{othmani2014tree}.

Finally, the current perching strategy is not designed for overhanging. We implemented (iii) to check that the perch site would result in a drone that was at most vertical with respect to the ground by checking the depth profile of the tree.

After filtering, the detector then determines the centroid of the tree trunk and the corresponding depth to the perch site centroid. Depth estimates are also filtered for noise. Using the depth and the image space centroid, along with the intrinsic and extrinsic parameters of the camera, we can then determine the pose of the target perch site. 

\subsubsection{Trajectory Planner}

\begin{figure}[!t]
\centering
\includegraphics[width=\columnwidth]{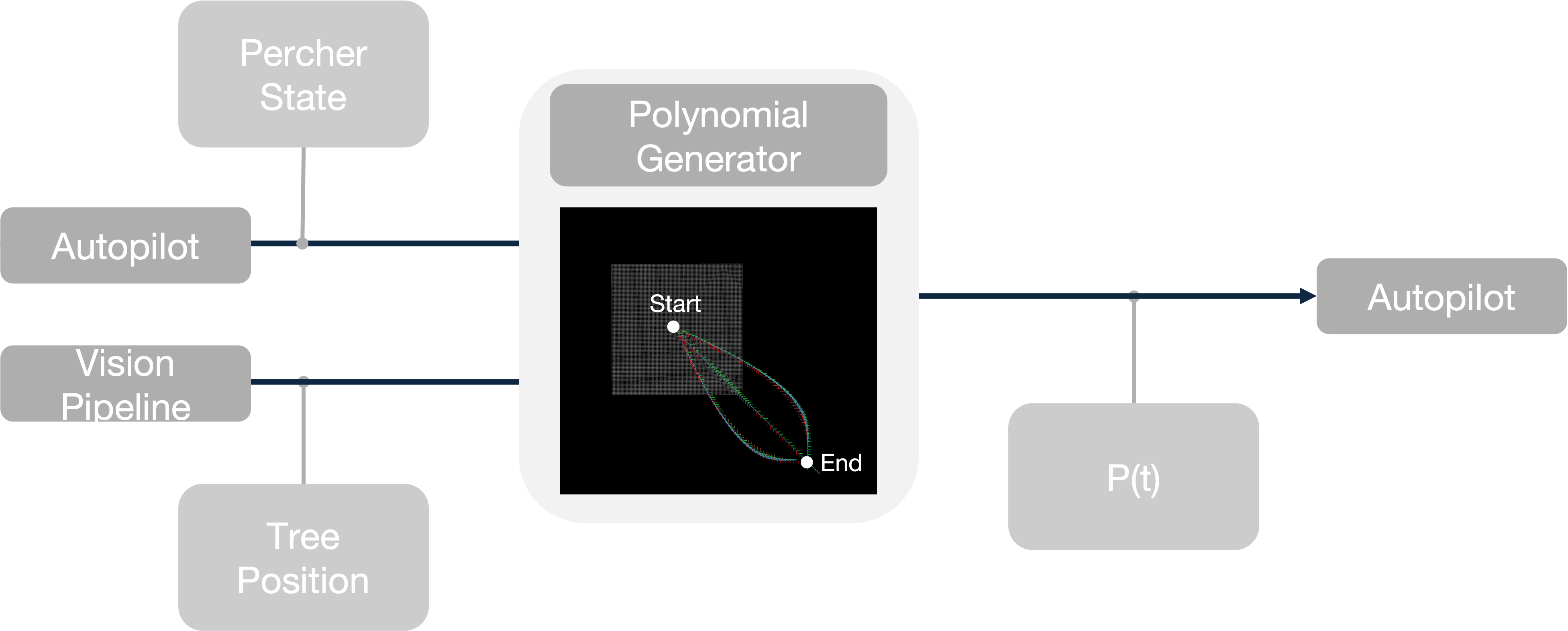}
\caption{Planner pipeline overview. The planner takes in tree state and drone state to generate polynomial trajectories to the target. The published outputs include the trajectory $P(t)$ and setpoints for the lower-level controller to follow.}
\label{fig:plannerpipeline}
\end{figure}

To reach the target perch site, the drone needs a perch trajectory $P(t)$ sampled in time. The planner takes as input both the current state of the drone and the estimated state of the tree, and outputs a time-based differential polynomial trajectory that satisfies position and velocity constraints. From this trajectory, the controller samples both position and velocity setpoints at discrete times for execution. 

The perching planner, shown in Fig.~\ref{fig:plannerpipeline}, processes the inputs through a simple polynomial generator with constraints on higher-order derivatives to ensure dynamically feasible motion. Important constraints for the trajectory include the current position of the perching drone and a small velocity, on the order of \SI{0.1}{\meter\per\sec}, at the desired perch site, normal to and into the perch site. This is necessary as it assists with engaging the grasping mechanism and ensures some forward velocity into the tree rather than stopping before the tree. The implication of this is a more secure grasp, however potentially leads to more stress on the gripper structures. Once the drone successfully reaches the perch location, the system transitions into the perching sequence and activates an IMU-based perch failure detection routine as described in the next session. 

\subsubsection{IMU-based perching failure detection}

A valuable contribution in this work is autonomous perch failure detection, which provides a critical safeguard for larger UAVs where failed attachment could result in severe structural damage. Because microspine adhesion is inherently stochastic, unexpected detachment is a concern for any microspine-based robotics system.

In our system, the planner continually monitors the magnitude of the drone's acceleration. We exploit the fact that in free fall, we would expect the accelerometer to read \SI{0}{\meter\squared\per\sec}. Therefore, if the drone begins to slip, we expect the acceleration to decrease from approximately \SI{9.8}{\meter\squared\per\sec} in hover at the beginning of the perching attempt, towards \SI{0}{\meter\squared\per\sec}. 

We define a threshold of \SI{7.0}{\meter\per\sec\squared} through through 5 trials, and empirically determine that our system latency from accelerometer to motor command is about \SI{100}{\milli\second}. If the acceleration decreases below the given threshold, the planner commands the drone to a safe position offset \SI{1}{\meter} away from the perch site, in the opposite direction of the tree. This approach successfully enables recovery from slipping during perching within \SI{100}{\milli\second}.  

The displacement during free fall is given by
\begin{equation}
    d = \tfrac{1}{2} g t^2,
\end{equation}
where $g = 9.8~\si{\meter\per\squared\second}$ is the gravitational acceleration 
and $t$ is the elapsed time. For $t = 0.1~\si{\second}$, we expect a fall of about \SI{5}{\centi\meter} on our system before the propellers could begin to respond.

\section{Waterfall System Design Process}

\begin{figure}[!t]
\centering
\includegraphics[width=0.7\columnwidth]{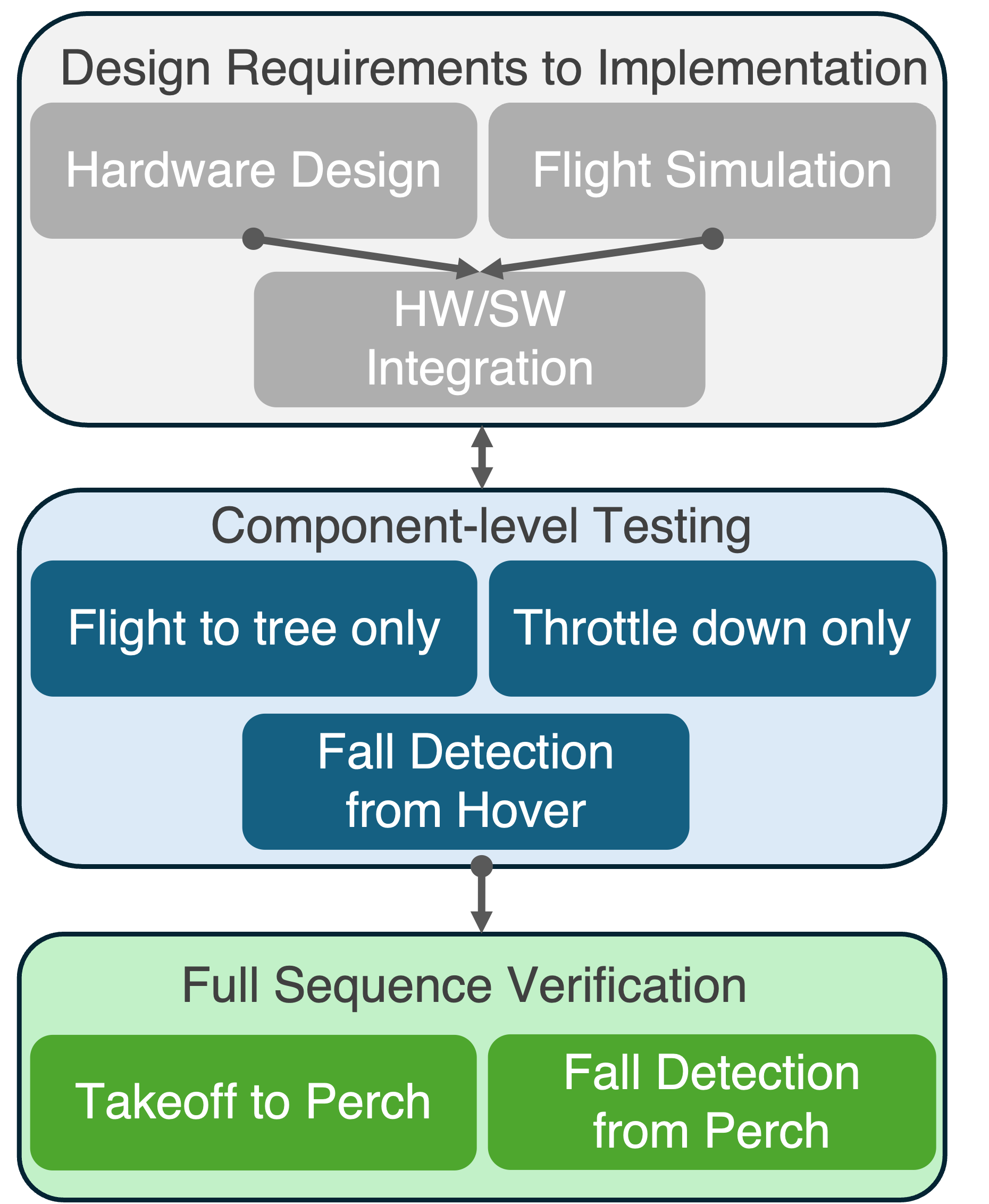}
\caption{Waterfall system design and test process. In our design process, we first discussed design requirements, and leaned on flight simulation to test out algorithms in parallel to hardware development. Then we moved to component-level testing, before finally validating the full perch sequence and failure recovery.}
\label{fig:designprocess}
\end{figure}

The development of the perching system followed a structured waterfall design process, illustrated in Fig.~\ref{fig:designprocess}. At each stage, design requirements were translated into testable hypotheses, with outputs feeding directly into the next phase of development.

In the requirements phase, we first defined the operational constraints of the system and made key design choices---perching on vertical tree trunks with a \SI{1.2}{\kilo\gram} quadrotor by minimizing impact energy, enabling recovery from perch failure, and ensuring autonomous perception and planning. These requirements shaped the key design tradeoffs, such as choosing an active gripper to relax velocity requirements versus the additional weight incurred from electronics. 

In parallel, we implemented a high-fidelity environment in Gazebo integrated with ROS-1 and the PX4 autopilot to allow simultaneous development of the perception and planning subsystems while hardware design was underway. The planning and perching environment is simulated using ROS 1 to integrate the PX4 autopilot and the Gazebo physics-based simulation environment. In this environment, both the vision and planning algorithms were thoroughly validated before moving onto indoor flight experiments. A textured cylindrical model of a tree trunk was included to test the full vision pipeline, in addition to the motion planning and flight control components of the robot. Because the simulated pipeline mirrored the real system architecture, using this simulation environment minimized the recalibration effort needed when transitioning the algorithms for the real flight environment.

In the component testing phase, individual subsystems were fabricated and evaluated independently. For instance, the standalone gripper was tested on benchtop and on trees to characterize closure speed, spine adhesion, and compliance under misaligned impacts. The gentle perch controller was tuned on a pivot-mounted rail testbed, allowing safe iteration on thrust decay profiles. 

The final integration and validation phase assembled the complete system on the Uvify IFO-S quadrotor and progressed to flight trials in the indoor flight room. At this stage, the full hardware and software system was tested under human-in-the-loop supervision, including the vision-based perch site detection, trajectory planning, gripper triggering, and failure recovery. The stepwise progression of the waterfall process ensured that risks were addressed early, failures were contained within subsystems, and the final demonstrations were built on validated components.

\section{Flight Validation and Results}

All indoor flights were performed in the Boeing Flight \& Autonomy Laboratory at Stanford University. An NVIDIA Orin was used for off-board processing, with information passed to and from the drone using ROS 1.  The quadrotor used in this work is a modified Uvify IFO-S quadrotor. The stock quadrotor is \SI{1}{\kilo\gram} at takeoff, including the battery and propellers (Size 7042 plastic propellers). With the perching hardware included, the final quadrotor is \SI{1.2}{\kilo\gram} at takeoff. 

The drone's onboard computer is an NVIDIA Jetson Nano which interfaces with the low level PX4 autopilot and the high level ROS environment. The modified quadrotor was fitted with a 3D-printed bumper (black PLA printed using a Prusa MK3S+) to gently rest the bottom against a surface during perching. A foam pool noodle was adhered to the bumper for cushioning. Each leg is also fitted with foam balls, adhered with double-sided tape, in order to facilitate soft landings during testing. The perching mechanism assembly is attached onto the quadrotor with standard metric screws and adds about \SI{200}{\gram} of weight.

\subsection{Subsystem Validation of the Gentle Perch Controller}

\begin{figure}[!th]
\centering
\includegraphics[width=0.7\columnwidth]{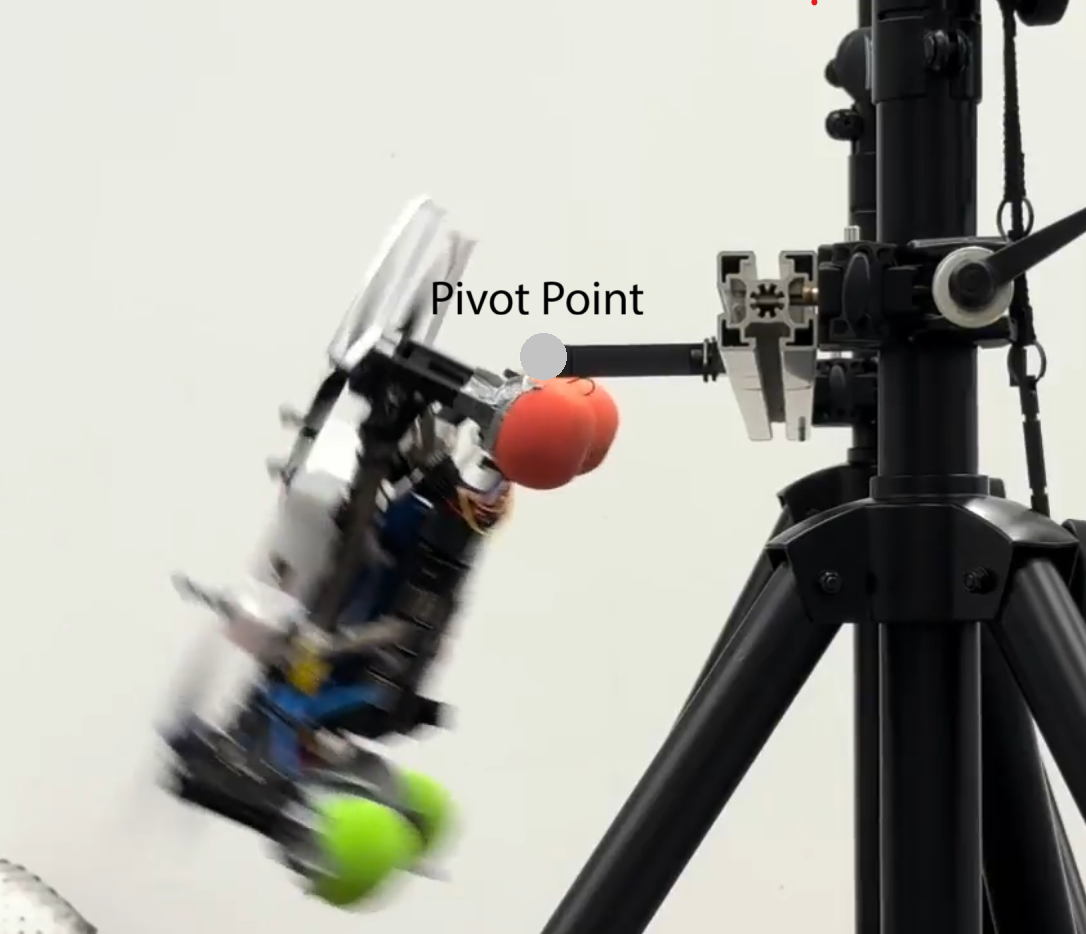}
\caption{Gentle Perching Controller Test Fixture.}
\label{fig:gentleperch}
\end{figure}
To safely tune the throttle-down maneuver, we built a throttle-down test apparatus consisting of a \SI{1}{\meter} length 80/20 stock rail on two stable height-aligned tripods, as shown in Fig.~\ref{fig:gentleperch}. The quadrotor is then rigidly bolted to the rail with a $90^\circ$ pin joint, allowing the robot to swing as the hover thrust is decreased. This experimental setup mimics the one degree of freedom conditions during perching and allows for controller tuning with minimal risk to the drone and eliminating extra variables. 

\subsection{Human-in-the-Loop Autonomous Flight Experiments}

\begin{figure}[!t]
\centering
\includegraphics[width=\columnwidth]{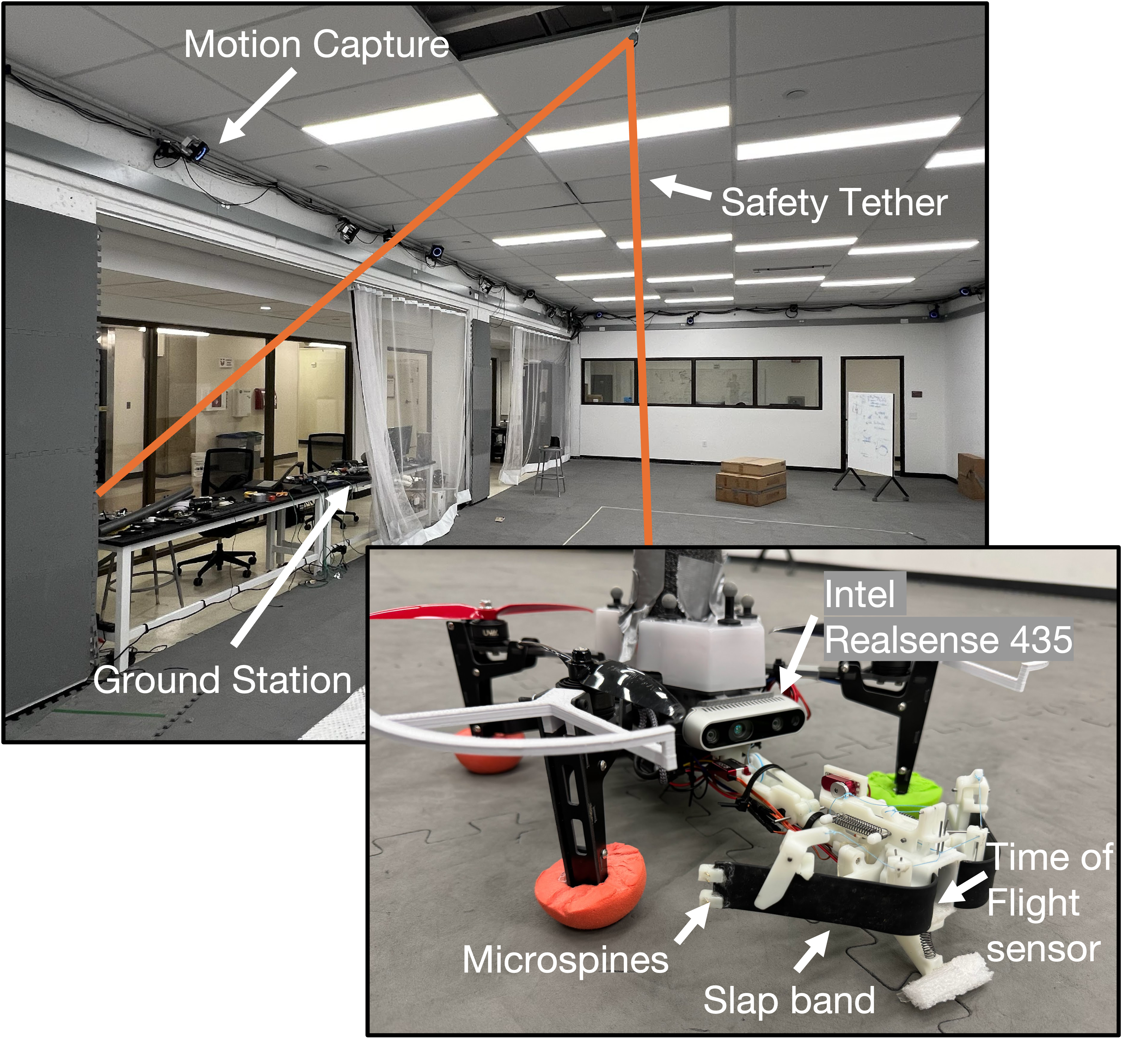}
\caption{Indoor flights were conducted in a motion capture environment with a safety tether (outlined in orange) attached to the quadrotor. The Uvify IFO-S quadrotor was modified with the perching hardware for flight experiments.}
\label{fig:flightroom_and_hardware}
\end{figure}

To ensure safety to both the robot and human operators, the flight room floor was lined with foam and the quadrotor was attached to a safety tether hooked through a pulley at the ceiling. The tether was not in tension for any flight tests, but in the event of an unplanned accident, the tether could have been pulled to prevent a crash. The flight room setup is discussed in more detail in Appendix~\ref{appendix:flightroomsetup}.

The possible failure modes of our system include: (i) failure to detect tree, (ii) failure to localize tree, (iii) failure to plan an effective trajectory,  (iv) failure to trigger,  (v) failure to grasp, (vi) slip during grasping/perching, (vii) slip after perching, and (viii) failure to recover from slip.

\begin{figure}[!t]
\centering
\includegraphics[width=\columnwidth]{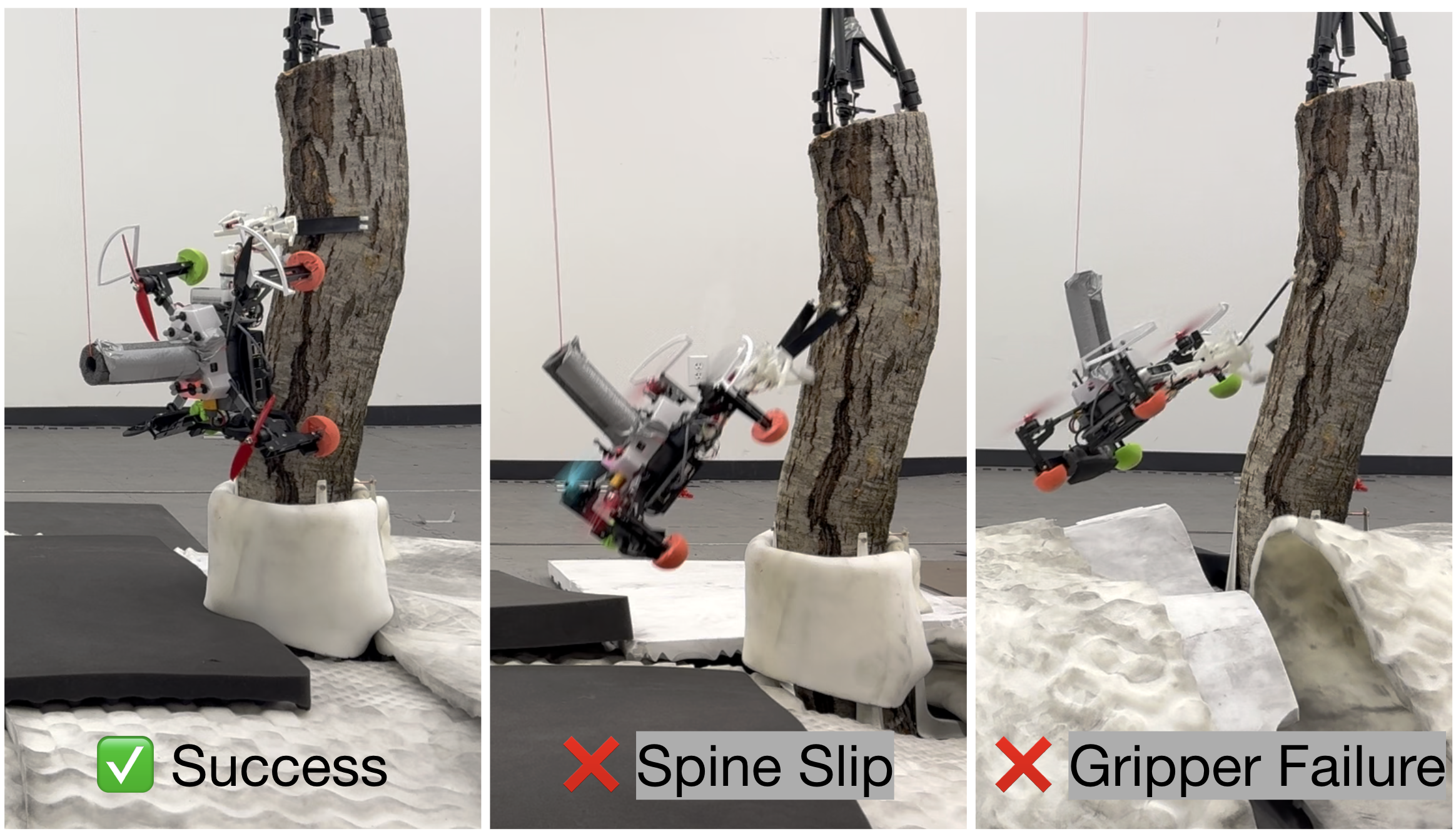}
\caption{Stills from the flight experiments are shown from left to right: (i) an example of the end state for perch success, (ii) free fall during the spine slip failure case, and (iii) free fall during gripper failure case (in this instance, the slapbands not adhered properly). }
\label{fig:flightexperiment}
\end{figure}

Across a total of 20 human-in-the-loop autonomous indoor flight trials on an oak tree segment, the system achieves a \textbf{75\% perching success rate}. A successful trial is defined as the drone detecting the tree perch site, generating a correct trajectory, reaching the perch site, triggering the gripper, and maintaining attachment for at least \SI{10}{\second} without slipping. The end state of a successful perch is shown in Fig.~\ref{fig:flightexperiment}.

Of the five failed trials, two were due to the microspines stochastically slipping off before the \SI{10}{\second} mark after a would-be-successful perch, and three were gripper failures. We further broke down the gripper failure cases as two were attributed to the slap bands slipping out of their holders due to insufficient glue while prototyping, and one was due to a surprise catastrophic failure of the gripper plastic. These failures are shown in Fig.~\ref{fig:flightexperiment}. These failure modes suggest that with more robust gripper prototyping, the autonomy success rate would be higher. Vision failure modes are caught earlier on by having the human in the loop, and further investigation is needed to identify the shortcomings of the perception system on trees.

To evaluate perch failure recovery, we induced controlled failures in 2 additional flights by intentionally filing down the microspines of the gripper so that they were dull and unlikely to fully engage. We then ran the same autonomous perching experiments. In both cases, the IMU-based failure detection correctly identified the onset of slipping within \SI{100}{\milli\second}, triggering the recovery maneuver with a \textbf{100\% perch failure recovery success rate}. The drone stabilized at a commanded offset of \SI{1}{\meter} away from the tree, providing preliminary validation of an accelerometer-based method for failure detection in larger UAVs.

More videos and code of flight tests can be found at our project website: \url{website-released-on-acceptance}.

\section{Conclusion and Future Work}

This work presents a preliminary investigation of a gentle perching and perch failure recovery approach for payload-bearing aerial robots on vertical surfaces. Using our knowledge of the sufficiency region in velocity space for passive grippers, we made the design choice to develop a compliant active gripper to minimize the relative velocity necessary at impact to trigger the gripper. A vision-based perch site detector, trajectory planner, and visual tracking pipeline were combined with an IMU-based failure detection and reaction routine to realize a complete autonomous perching framework. The system was validated in human-in-the-loop autonomous indoor experiments on a real oak tree, achieving a 75\% perching success rate across 20 trials and a 100\% recovery rate across 2 induced perch failures.

There are many interesting avenues to investigate beyond this work. First, the gripper mechanism does not necessarily need to be active; a completely passive gripper solution could exist that both minimizes the response time and weight without requiring a high relative velocity to trigger. Changing to a passive gripper system would leave more room in the mass budget to carrying scientific instruments or additional onboard processing. Furthermore, it would be interesting to investigate ways to optimize the contact of the microspines on the tree bark surface via load sharing. 

Second, the trajectory planner in this study employed a polynomial formulation for simplicity; extending to more advanced trajectory frameworks could improve reliability and safety in cluttered forest environments.

Finally, while the motion capture system was used in experiments to provide ground truth localization, future iterations of this platform should integrate onboard SLAM or VIO systems to enable fully autonomous perching in unstructured outdoor settings. Together, these directions point toward robust, scalable perching systems that could expand the operational capabilities of UAVs for ecological monitoring, infrastructure inspection, and other long-duration aerial tasks.


\acknowledgements 
The authors thank James Biddle and the University Tree Program for permission to use a segment of a real tree for indoor flight testing. The authors also thank Jun En Low for his instructions on drone piloting and Joshua Dong for his work in early prototyping of the gripper design. This work was supported by EpiSys Science Inc. and subcontracted to Stanford’s Center for Design Research. J. Di was additionally supported by a P.E.O. Scholarship and the Zonta International Amelia Earhart Scholarship at the time of work. In addition, J. Di, K. A. W. Hoffmann, and T. G. Chen were with the Department of Mechanical Engineering at Stanford University at the time of work.

{\appendices
\input{appendices}}

\bibliographystyle{IEEEtran}
\bibliography{perching}

\thebiography
\begin{biographywithpic_square}
{Julia Di}{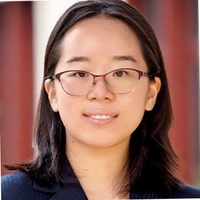} received the B.S. degree in electrical engineering and computer science from Columbia University, and received the M.S. and Ph.D. degrees in mechanical engineering from Stanford University, Stanford, CA in 2020 and 2025, respectively. \\
She was with the Department of Mechanical Engineering, Stanford University, Stanford, CA at the time of work. Her research interests include tactile sensing, autonomy, and grasp perception. 
\end{biographywithpic_square} 

\begin{biographywithpic_square}
{Kenneth A. Hoffmann}{fig/biography/hoffmann}
received the B.S. degree in mechanical engineering from the University of Illinois Urbana–Champaign, Urbana, IL, USA, in 2017, and the M.S. and Ph.D. degrees in mechanical engineering from Stanford University, Stanford, CA, USA, in 2019 and 2023, respectively. \\
He was with the Department of Mechanical Engineering, Stanford University, Stanford, CA at the time of work. His doctoral research focused on design principles and systems analysis for visually guided aerial grasping robots. His research interests include robotics, mechatronics, design, controls, and autonomy, with applications in aerial robotics.
\end{biographywithpic_square}

\begin{biographywithpic_square}
{Tony G. Chen}{fig/biography/chen}
received the B.S. degree in mechanical engineering from the Georgia Institute of Technology, Atlanta, GA, USA, and the M.S. and Ph.D. degrees in mechanical engineering from Stanford University, Stanford, CA, USA, in 2020 and 2023, respectively. \\
He was with the Department of Mechanical Engineering, Stanford University, Stanford, CA at the time of work. He is now an Assistant Professor with the George W. Woodruff School of Mechanical Engineering, Georgia Institute of Technology. His research interests include bio-inspired and field robotics, particularly the design of climbing and perching robots that interact with challenging, real-world environments.
\end{biographywithpic_square}

\begin{biographywithpic_square}
{Tian-Ao Ren}{fig/biography/ren}
received the B.Eng. degree in robotics engineering from Beijing University of Chemical Technology, Beijing, China, in 2023, and the M.S. degree in mechanical engineering from Stanford University, Stanford, CA, USA, in 2025, where he is currently pursuing the Ph.D. degree in mechanical engineering. \\
His research interests include medical robots, tactile sensing, soft robots, and sim-to-real applications.
\end{biographywithpic_square}

\begin{biographywithpic_square}
{Mark R. Cutkosky}{fig/biography/cutkosky} received the Ph.D. degree in mechanical engineering from Carnegie Mellon University, Pittsburgh, PA, USA, in 1985. \\
He is currently the Fletcher Jones Professor of Mechanical Engineering with Stanford University, Stanford, CA, USA. His research interests include bioinspired robots, haptics, and rapid prototyping processes. \\
Dr. Cutkosky is a Fellow of IEEE, ASME, and an IEEE RA-L Pioneer in robotics.
\end{biographywithpic_square}

\end{document}

%% file: 1_preamble.tex
\usepackage{amsmath,amsfonts}
\usepackage{algorithmic}
\usepackage{algorithm}
\usepackage{array}
\usepackage{caption}
\usepackage[caption=false,font=normalsize,labelfont=sf,textfont=sf]{subfig}
\usepackage[dvipsnames]{xcolor}
\usepackage{textcomp}
\usepackage{stfloats}
\usepackage{url}
\usepackage{verbatim}
\usepackage{graphicx}
\usepackage{cite}
\usepackage{cleveref}
\usepackage{siunitx}
\newcommand{\jd}[1]{\textcolor{OliveGreen}{(JD: #1)}}

%% file: appendices.tex
\section{FEA for Microspine Placement}
\label{appendix:gripperfea}

\begin{figure}
    \centering
    \includegraphics[width=\linewidth]{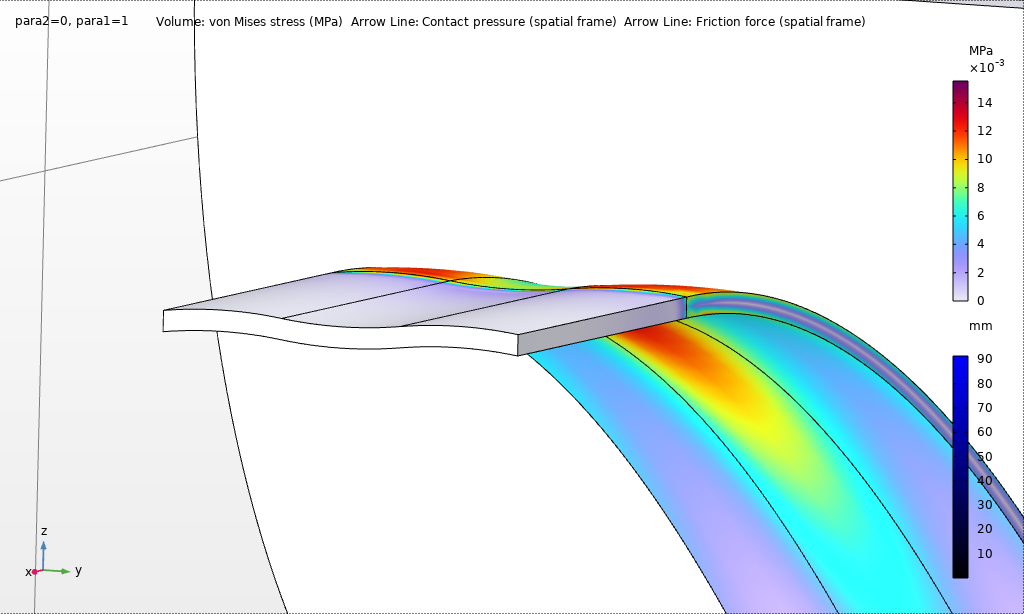}
    \caption{FEA plot showing that contact stress is greatest at the tip of the band for an elastic band wrapped around a curved object.}
    \label{fig:fea}
\end{figure}

A preliminary simulation was conducted in COMSOL Multiphysics (v6.2) to investigate the contact stress of one of the bistable bands when engaged onto a cylindrical surface, as shown in Fig.~\ref{fig:fea}. This simulation was performed with two stationary steps. The first step involves movement in two directions to position the pre-bending gripper into a ``snapped open" state. The second step involves movement in one direction to simulate the latching motion. 

The gripper was modeled as a hyperelastic rubber using the Yeoh constitutive model, with material parameters defined as $\rho = 1100 \,\mathrm{kg/m^3}$, $c_{1} = 100 \,\mathrm{kPa}$, $c_{2} = 6 \,\mathrm{kPa}$, and $c_{3} = -30 \,\mathrm{Pa}$, to capture the nonlinear large-deformation behavior of elastomers. The target object, representing a tree stump, was simplified as a rigid cylindrical body. The contact interactions between the gripper and the rigid body were modeled through the solid mechanics interface with appropriate contact pairs. Boundary conditions included a fixed constraint at the gripper base along with roller and symmetry constraints. 

To ensure accurate resolution of deformation and stress distribution, the computational domain was discretized using a predefined ``Extremely Fine'' mesh, with element sizes ranging from $0.0367 \,\mathrm{mm}$ to $3.67 \,\mathrm{mm}$, a curvature factor of $0.2$, and a maximum element growth rate of $1.3$. The COMSOL simulation reveals that the contact stress is unevenly distributed across the band, with the majority at the tip. Therefore it is beneficial to place the microspines at the tip of the band, to help them stay engaged with the surface.

\section{Flight Experiments Setup}
\label{appendix:flightroomsetup}

\begin{figure}
    \centering
    \includegraphics[width=\linewidth]{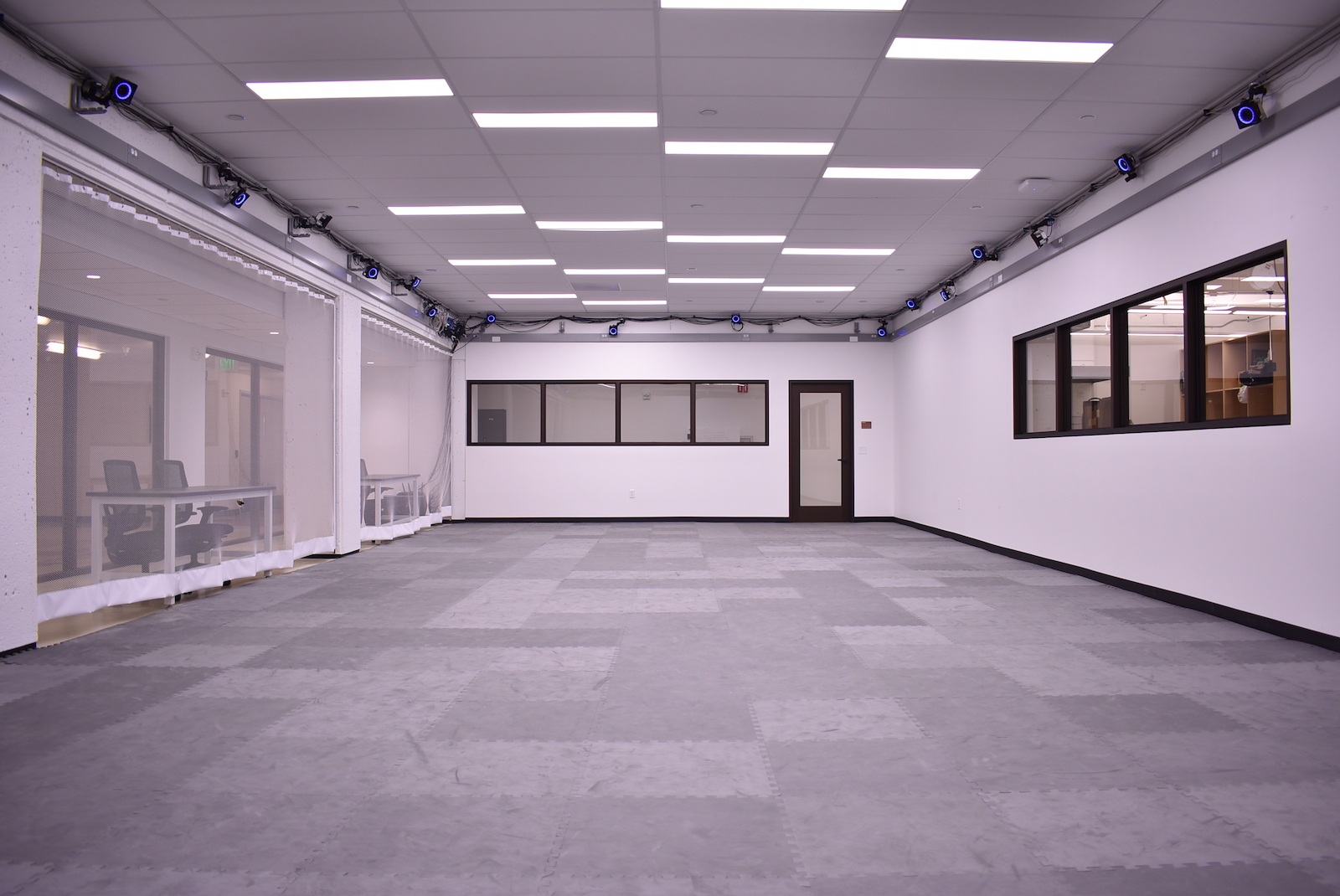}
    \caption{Stanford flight room setup.}
    \label{fig:flightroom}
\end{figure}

All indoor flights were conducted in Stanford’s Flight Room, a dedicated motion-capture facility as shown in Fig.~\ref{fig:flightroom}. The flight space is 16 × 6 × \SI{3}{\meter} equipped with 26 high-fidelity Opti-track infrared cameras for motion tracking. This provides sub-millimeter pose estimates, and offers a controlled, obstacle-free volume well suited for aerial robotics experiments. The environment supports tight safety protocols including a ceiling-mounted tether system for emergency intervention, padded flooring, and ample clearance in all axes.

\begin{figure}[!t]
    \centering
    \includegraphics[width=\linewidth, trim=8cm 4cm 4cm 20cm, clip]{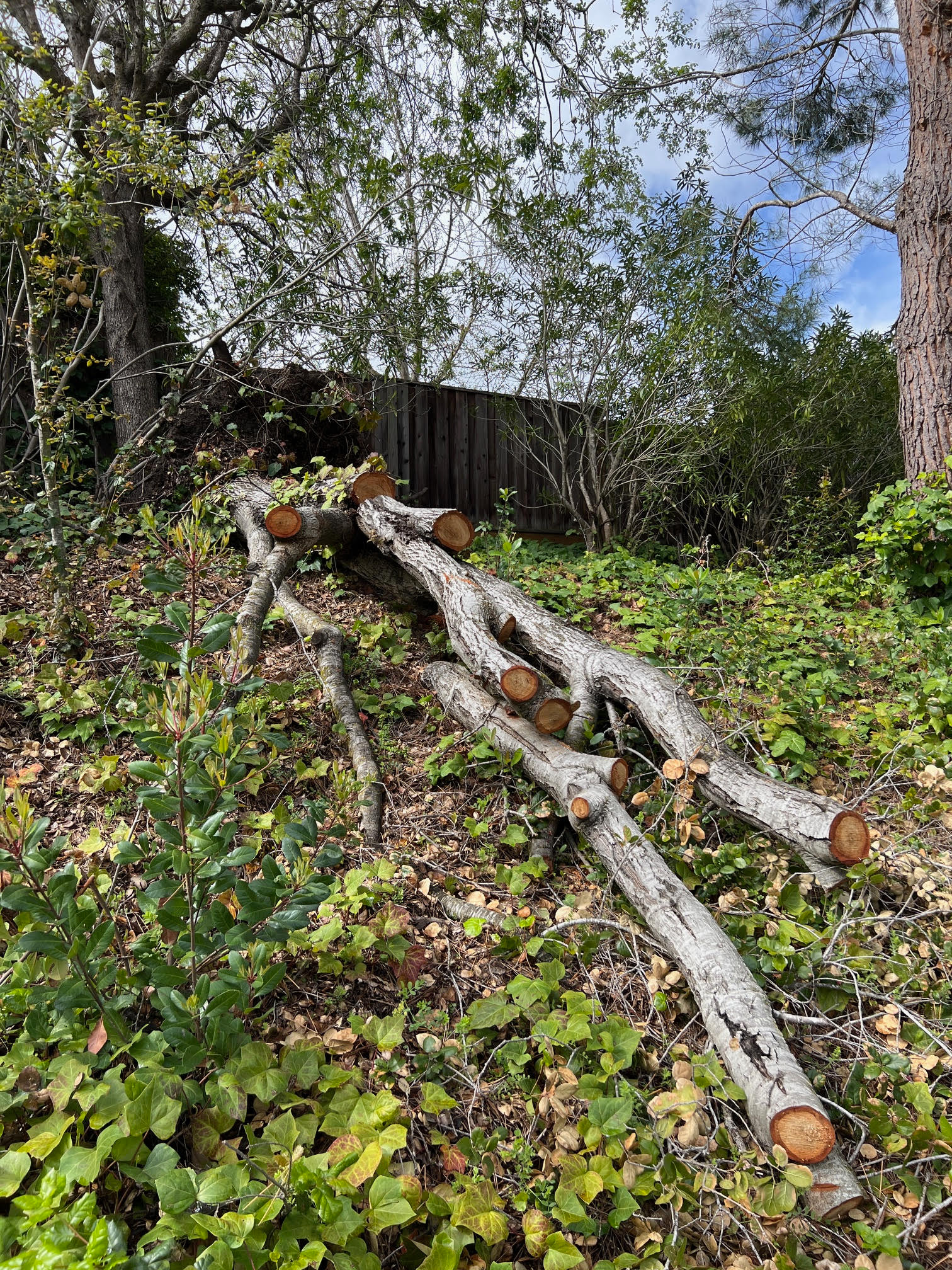}
    \caption{Oak tree segment obtained from a downed oak tree on Stanford's campus.}
    \label{fig:tree}
\end{figure}

For perching experiments, the tree trunk for testing was cut from a downed oak tree on campus shown in Fig.~\ref{fig:tree}, with permission from the university's Tree Program. The tree segment was manually sanded to ensure a flat bottom, then bolted to a stand with supports to remain stably upright during testing.

